\documentclass[10pt,twocolumn,letterpaper]{article}

\usepackage{iccv}      
\usepackage[linesnumbered,ruled,vlined]{algorithm2e}
\usepackage{bbding}
\usepackage[misc]{ifsym}

\definecolor{iccvblue}{rgb}{0.21,0.49,0.74}
\usepackage[pagebackref,breaklinks,colorlinks,allcolors=iccvblue]{hyperref}

\def\paperID{1866} 
\def\confName{ICCV}
\def\confYear{2025}

\def\METHODNAME{\textsc{AnyPortal}}
\def\METHODNAMEABS{AnyPortal} 

\title{\METHODNAME: Zero-Shot Consistent Video Background Replacement\vspace{-4mm}}

\author{Wenshuo Gao \hspace{12pt} Xicheng Lan  \hspace{12pt} Shuai Yang$^\textrm{\Envelope}$\\
\normalsize{Wangxuan Institute of Computer Technology, State Key Laboratory of Multimedia Information Processing,} \\\normalsize{Peking University, Beijing, China}\\
{\tt\small \{gaowenshuo, lanxicheng\}@stu.pku.edu.cn \hspace{12pt} williamyang@pku.edu.cn}\vspace{-4mm}
}

\begin{document}
\twocolumn[{%
\renewcommand\twocolumn[1][]{#1}%
\maketitle
\centering
\includegraphics[width=1.0\linewidth]{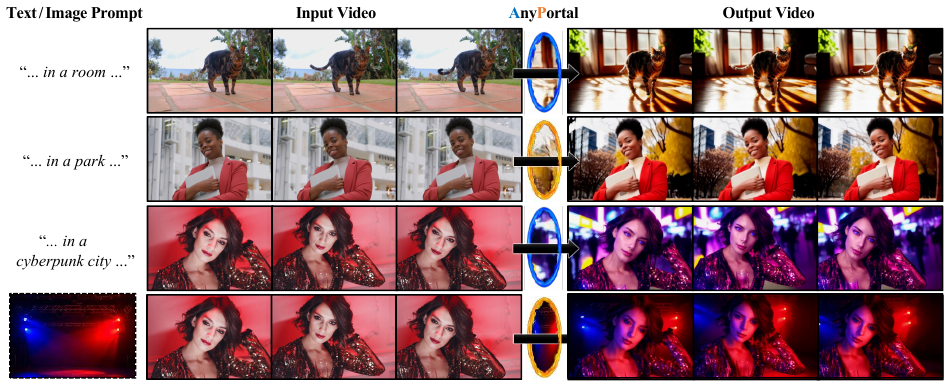}
 \vspace{-2em}
\captionof{figure}{\textbf{We propose \METHODNAME, a training-free framework for high-consistency video background replacement and foreground relighting.} Given an input foreground video and a text or image prompt of the background, our method produces a video with the target background under harmonious illuminations, while maintaining the foreground video details and intrinsic properties.\vspace{1em}}
\label{fig:teaser}
}]

\begin{abstract}
Despite the rapid advancements in video generation technology, creating high-quality videos that precisely align with user intentions remains a significant challenge. 
Existing methods often fail to achieve fine-grained control over video details, limiting their practical applicability. 
We introduce \textbf{\METHODNAMEABS}, a novel zero-shot framework for video background replacement that leverages pre-trained diffusion models. 
Our framework collaboratively integrates the temporal prior of video diffusion models with the relighting capabilities of image diffusion models in a zero-shot setting. 
To address the critical challenge of foreground consistency, we propose a Refinement Projection Algorithm, which enables pixel-level detail manipulation to ensure precise foreground preservation. 
\METHODNAMEABS~is training-free and overcomes the challenges of achieving foreground consistency and temporally coherent relighting. 
Experimental results demonstrate that \METHODNAMEABS~achieves high-quality results on consumer-grade GPUs, offering a practical and efficient solution for video content creation and editing.
\end{abstract}    
\section{Introduction}
\label{sec:intro}


Teleportation, often regarded as one of the most popular superpowers, offers the fascinating ability to travel anywhere instantly. While true teleportation remains confined to the realm of science fiction, digital technologies have made its virtual counterpart a reality, particularly in the film and entertainment industry. Through the use of green screens and digital techniques, actors can be seamlessly transported from a studio to virtually any location. However, this process is far from trivial. It involves a complex pipeline that includes constructing green screen environments, generating backgrounds that are geometrically consistent with the camera's perspective, and meticulously replacing the green screen with the synthesized background while ensuring realistic illuminations. Despite its widespread use in professional settings, this workflow remains resource-intensive and labor-expensive beyond normal users. 

%

Recent years have witnessed rapid advancement in AIGC, highlighting the potential to make ``virtual teleportation'' accessible to the general public. The state-of-the-art image diffusion model, IC-Light~\cite{zhang2025scaling}, enables users to replace the background of a photo with harmonized illuminations, achieving robust performance through extensive training on paired image datasets.
However, collecting large-scale paired video datasets is considerably more difficult compared to paired images, making scaling this approach to video significant challenges.
Meanwhile, recent cutting-edge video diffusion models~\cite{yang2024cogvideox, zheng2024open} demonstrate impressive capabilities in video generation and editing. 
Despite their potential, these models still fall short for widespread video background replacement tasks. First, existing video diffusion models exhibit limited controllability over generated content. While some approaches~\cite{CogvideoXControlnet, gu2025diffusion,shao2025finephys} introduce coarse controls for edges, poses and motion, they lack pixel-level precision, often resulting in unintended alterations to the foreground appearance. Second, adapting video models to our specialized task typically requires task-specific training or fine-tuning, which is hindered by the scarcity of paired video data and the substantial computational resources needed to train large video models.

We believe that pre-trained large diffusion models inherently possess rich prior knowledge for video background replacement: IC-Light provides valuable insight into how lighting should be rendered, while video models capture real-world dynamics. Our key insight is to explore \textit{to which extent these pre-trained models can manage tasks that extend beyond their original training task, collaboratively leveraging their inherent priors in a zero-shot setting.} 

To this end, we investigate the zero-shot video background replacement problem. 
While IC-Light excels at illumination harmonization, and video models provide powerful temporal priors, naively combining them fails to address the critical challenge of foreground consistency, which requires precise pixel-level control over the generation process.
%
While mature training-free control schemes exist for image models -- such as inference-time optimization~\cite{yu2023freedom, ye2024tfg} or DDIM inversion~\cite{song2020denoising} with latent manipulations~\cite{gao2024fbsdiff, lugmayr2022repaint} -- these methods face significant limitations when applied to video models:~1) Optimization on video models incurs prohibitive computational costs; 2) Video models typically operate in a highly compact 3D latent space~\cite{yang2024cogvideox}, which degrades inversion quality and hinders detailed manipulations. 
To address these challenges, we propose a novel \textbf{Refinement Projection Algorithm (RPA)} tailored for video models. RPA computes a projection direction in the latent space that simultaneously ensures high consistency with the input foreground details and high-quality background, offering a robust and efficient solution for zero-shot video background replacement.

We introduce \textbf{\METHODNAME}, a novel training-free framework for video background replacement. \METHODNAME~first generates a coarse video with illumination harmonized by IC-Light and then enhances its temporal consistency using a pre-trained video diffusion model. To achieve precise control over foreground details, a Refinement Projection Algorithm is proposed to enable pixel-level manipulation. As shown in Fig.~\ref{fig:teaser}, \METHODNAME~seamlessly transfers foreground subjects (\eg, humans or objects) from an input video to a new environment, specified by either a text prompt or a background image, while ensuring natural illuminations, realizing ``virtual teleportation'' in videos. Remarkably, our framework operates efficiently on a single 24GB GPU. Furthermore, its modular design allows each component to be implemented using the best available pre-trained models, ensuring compatibility with the latest advancements in AIGC.
Our contributions are threefold:
\begin{itemize}
    \item We introduce \METHODNAME, an efficient and training-free framework for video background replacement.
    \item We design a modular pipeline that integrates the latest pre-trained image and video diffusion models, to combine their strengths for realistic and coherent video generation.
    \item We propose a novel Refinement Projection Algorithm that enables pixel-level detail manipulation in compact latent spaces, ensuring precise foreground preservation.    
\end{itemize}


\section{Related Work}
\label{sec:related}

\textbf{Image Diffusion Model.} Latent Diffusion Model~(LDM) has become a strong method for image generation, notably gaining significant popularity with Stable Diffusion~\cite{rombach2022high}. 
The main idea is to first compress the image data into latent space using a Variational Autoencoder (VAE), then progressively denoise Gaussian noises with algorithms such as DDPM~\cite{ho2020denoising} and DDIM~\cite{song2020denoising}, and finally decode the denoised latent back to an image. 
Traditionally, U-Net architectures were used, followed by DiT \cite{peebles2023scalable} that introduces Transformers to improve generated results as in SD3~\cite{esser2024scaling}.

To meet user demands for generating images with specific conditions, 
SDEdit \cite{meng2021sdedit} allows for training-free image editing in the LDM framework by denoising the noisy image from an intermediate timestep.
Methods like ControlNet~\cite{zhang2023adding}, T2I-Adapter~\cite{mou2024t2i} and ControlNeXt~\cite{peng2024controlnext} create a learnable branch of the denoising model to offer additional control conditions such as edges and depth maps. 
Another approach to image editing is using DDIM inversion~\cite{song2020denoising} and Null-Text Inversion~\cite{mokady2023null}, which inverts the denoising process of a given image, then re-denoises it with text guidance and attention manipulations~\cite{hertz2022prompt,parmar2023zero,tumanyan2023plug,cao2023masactrl}.

\begin{figure*}[tp]
    \centering
    \includegraphics[width=1\textwidth]{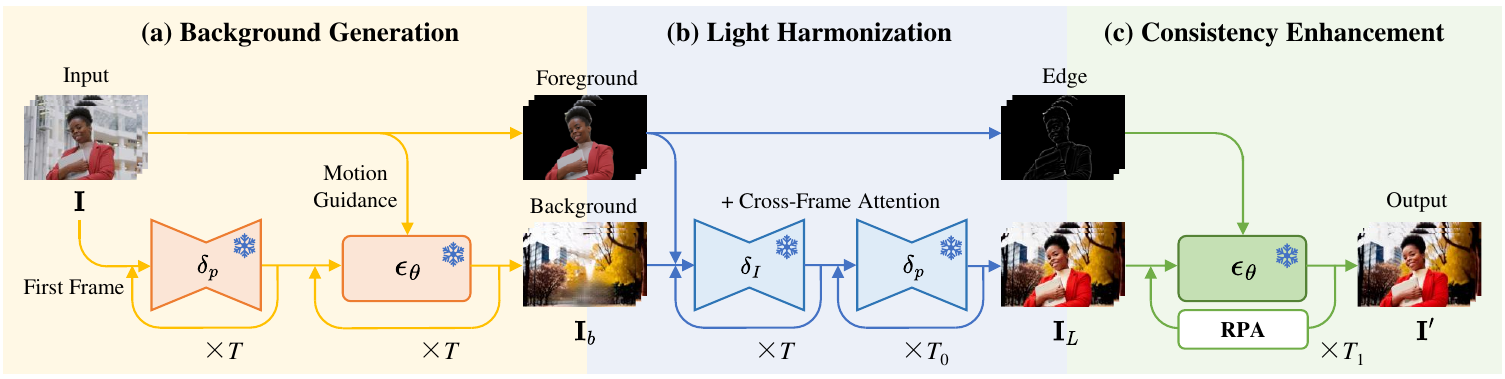}\vspace{-2mm}
    \caption{\textbf{Framework of \METHODNAME.} (a) Background Generation: A video diffusion model $\epsilon_\theta$ is used to generate a basic background video $\mathbf{I}_b$ following the first frame generated by IC-Light model $\delta_p$ and the camera motion of the input video $\mathbf{I}$; (b) Light Harmonization: a two-step pipeline based on IC-Light model $\delta_I$ and $\delta_p$ is proposed to combine the foreground and background video and harmonize its illumination; (c) Consistency Enhancement: The video diffusion model $\epsilon_\theta$ is used to improve the temporal consistency of $\mathbf{I}_L$, with a novel Refinement Projection Algorithm~(RPA) to further strengthen the foreground consistency with the input $\mathbf{I}$.  (\SnowflakeChevron: All models are frozen)}\vspace{-4mm}
    \label{fig:pipeline}
\end{figure*}

\textbf{Video Diffusion Model.}
Many works attempt to extend image diffusion models to video diffusion generation. 
Early attempts make partial modifications to image models, by redesigning the sampling scheme for zero-shot video generation~\cite{khachatryan2023text2video}, fine-tuning inflated models for one-shot video generation~\cite{wu2023tune}, and training a plug-and-play temporal module to turn image models into animation generators~\cite{guo2023animatediff}. With increased computational resources, full model training on large-scale video dataset has been proposed~\cite{ho2022imagen,singer2022make,he2022latent,zhou2022magicvideo,blattmann2023align,blattmann2023stable}. 
Modern practices such as OpenSora~\cite{zheng2024open} and CogVideoX~\cite{yang2024cogvideox} have focused on learning in the 3D latent space~(2D for spatial + 1D for temporal). These methods typically extend the 2D image VAE to 3D VAE that compresses both spatial and temporal dimensions, mapping videos into a 3D latent space. The denoising model works in this latent space, usually using a DiT architecture~\cite{peebles2023scalable}, which exhibits better temporal consistency and scalability.


Diffusion models can be used for video editing. 
An intuitive idea is to apply image editing techniques like SDEdit~\cite{meng2021sdedit} and Prompt-to-Prompt~\cite{hertz2022prompt} to the image models with cross-frame attention to strengthen temporal consistency~\cite{qi2023fatezero,zhang2023controlvideo,yang2023rerender,yang2024fresco,geyer2023tokenflow}. However, image models intrinsically lack modeling of real-world motion, leading to unnatural dynamics. 
Meanwhile, due to the high complexity and compactness of 3D latent space, the above editing and DDIM inversion techniques~\cite{song2020denoising} are not directly compatible with the DiT-based diffusion models, resulting in motion degradation and appearance distortions.  
To leverage the latest advancements of video models, we propose an effective Refinement Projection Algorithm to maintain the input video details without harming the generated motions.


\textbf{Foreground Relighting and Background Replacement.} TotalRelighting~\cite{pandey2021total} and SwitchLight~\cite{kim2024switchlight} train neural networks to predict  surface normals and albedo to recompute new lighting. Relightful Harmonization~\cite{ren2024relightful} fine-tunes an image diffusion model conditioned on the background for foreground relighting.
IC-Light~\cite{zhang2025scaling} simultaneously achieves impressive background replacement and illumination harmonization by concatenating the input noise and foreground condition (and optionally background condition) before feeding into the diffusion model and finetuning the model with light transport consistency. Currently, there are few diffusion models specifically for video background replacement and foreground relighting. RelightVid~\cite{fang2025relightvid} combines IC-Light and AnimateDiff~\cite{guo2023animatediff} with finetuning. 
By comparison, our method does not require any training, achieving high compatibility and modularity. Each of the modules can be implemented by the best pretrained models, allowing us to leverage the latest advancements (\eg, CogVideoX~\cite{yang2024cogvideox}) for better video consistency.


\section{\METHODNAME}
\label{sec:method}

\subsection{Preliminary}

\paragraph{Video Diffusion Model.}
The latest video diffusion generation models~\cite{yang2024cogvideox,zheng2024open} 
typically includes a 3D VAE $\mathcal{D}\circ\mathcal{E}$ and a denoising DiT model $\epsilon_\theta$. The VAE consists of a video encoder $\mathcal{E}$ to encode a video clip $\mathbf{I}$ into the compressed latent feature $x=\mathcal{E}(\mathbf{I})$
and a video decoder to the latent back into a video $\mathbf{I}=\mathcal{D}(x)$. 
%
The denoising DiT model $\epsilon_\theta$ is trained for denoising in the latent space. 
At the timestep $t$, $\epsilon_\theta$ takes as input noisy latent $x_t$ and condition $c$ (typically, the prompt) to output a noise prediction $\epsilon_{\theta}(x_t, c, t)$. 
A clean $x_0$ could be sampled from a Gaussian noise $x_T\sim \mathcal{N}(0,1)$ by iteratively predicting $x_{t-1}$ from $x_t$ following denoising schemes such as DDIM~\cite{song2020denoising}, 
\begin{equation}\label{eq:ddim}
x_{t-1} = \sqrt{\alpha_{t-1}} x_0^t + \sqrt{1 - \alpha_{t-1}} \epsilon_{\theta}(x_t, c, t),
\end{equation}
where $\alpha$s are a group of parameters related to $t$ and $x_0^t$ is the denoised latent at timestep $t$,
\begin{equation}\label{eq:x0t}
x_0^t = \frac{x_t - \sqrt{1 - \alpha_t} \epsilon_{\theta}(x_t, c, t)}{\sqrt{\alpha_t}}.
\end{equation}
Finally, the generated video $\mathbf{I}=\mathcal{D}(x_0)$ is obtained.

\paragraph{IC-Light.}
IC-Light~\cite{zhang2025scaling} is an image diffusion model for background replacement and foreground relighting. It has two versions $\delta_p$ and $\delta_I$. 
$\delta_p$ is conditioned on prompts $p$ describing the appearance of the background, while $\delta_I$ is conditioned on background images $I_b$.
Given a foreground image $I_f$, it generates $I'=\delta_p(I_f, p)$ or $I'=\delta_I(I_f, I_b)$ with the corresponding foreground and background under harmonized illumination. For simplicity, we omit the iterative sampling operations and transformations between the image space and the 2D latent space. We experimentally find that the text-guided model excels at intensively harmonized illumination, whereas the image-guided model can create more consistent background. We will detail how we combine the two models to leverage their strengths in Sec.~\ref{sec:method-2} .


\subsection{Zero-Shot Video Background Replacement}

As illustrated in Fig.~\ref{fig:pipeline}, our framework is divided into three stages: (1) Background Generation; (2) Light Harmonization; (3) Consistency Enhancement. 
Our input is a foreground video $\mathbf{I}$ and a prompt $p$ describing the background. In the first stage, we generate a background video $\mathbf{I}_b$ that matches the camera movements of $\mathbf{I}$ with the help of a pre-trained video diffusion model $\epsilon_\theta$. The second stage harmonizes the lighting of the foreground object in the new background based on our proposed two-step IC-Light pipeline to produce a coarse video $\mathbf{I}_L$. The third stage introduces a novel Refinement Project Algorithm (RPA) that addresses inconsistencies between frames and refines the foreground details to match those of $\mathbf{I}$, yielding the final video $\mathbf{I}'$.

Note that our method is zero-shot and modular, fully leveraging the powerful generative ability of the pre-trained diffusion models $\epsilon_\theta$ and $\delta$s without any training or inference-time optimization. This allows us to generate impressive videos on a single 24GB-memory GPU. Moreover, our method fully benefits from rapidly growing vision diffusion research, as it can be implemented on the latest pretrained models once available to boost the performance. 


\begin{figure}[t]
    \centering
    \includegraphics[width=\linewidth]{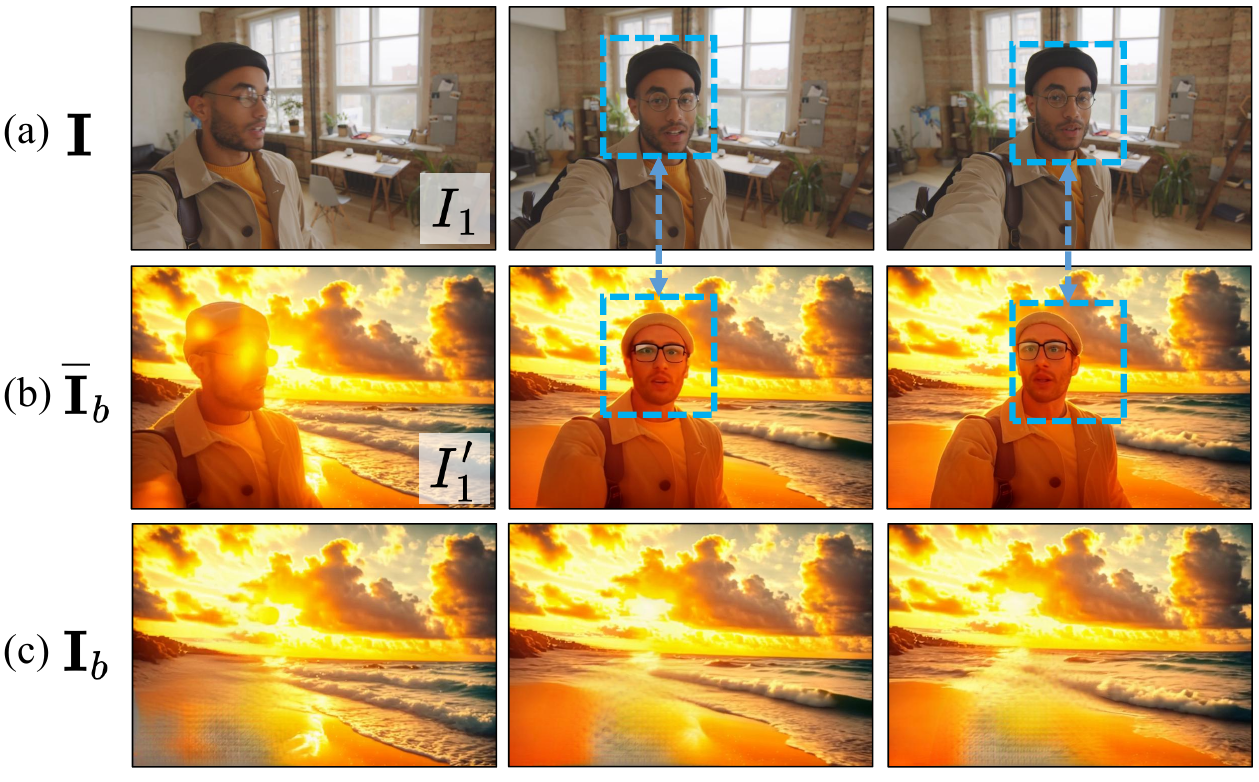}\vspace{-2mm}
    \caption{\textbf{Motion-aware background generation.} The generated background video $\mathbf{I}_b$ follows the camera motion of the input video. (a) Input video $\mathbf{I}$. (b) Video output $\mathbf{\overline{I}}_b$. (c) Inpainted output $\mathbf{I}_b$. The blue dotted lines indicate the inconsistent foreground areas.}\vspace{-4mm}
    \label{fig:bg}
\end{figure}

\subsubsection{Background Generation}


To seamlessly integrate the foreground with the background, the first stage produces a basic background video $\mathbf{I}_b$ that corresponds with the background prompt $p$ and, crucially, matches the camera motion of the $\mathbf{I}$.
To this end, we follow Diffusion-As-Shader~(DAS)~\cite{gu2025diffusion}, a ControlNet-based video generation framework that guides the video diffusion model with the first frame and the motion (tracked 3D points) of a guiding video. 
Specifically, $\mathbf{I}$ serves as the guiding video.
To obtain the first frame, we apply IC-Light to the first frame $I_1$ of $\mathbf{I}$, resulting in $I'_1=\delta_p(I_1, p)$. 
Then, we apply DAS to the backend video diffusion model to generate $\mathbf{\overline{I}}_b$ based on $I'_1$ and $\mathbf{I}$.
As shown in Fig.~\ref{fig:bg}, $\mathbf{\overline{I}}_b$ has the same camera motion as the input $\mathbf{I}$, but its foreground object may differ significantly from $\mathbf{I}$ and cannont be directly used as our video background replacement result. Finally, we use ProPainter~\cite{zhou2023propainter} to remove the foreground object, obtaining the basic background video $\mathbf{I}_b$.

\subsubsection{Light Harmonization}
\label{sec:method-2} 

\begin{figure}[t]
\centering
\includegraphics[width=\linewidth]{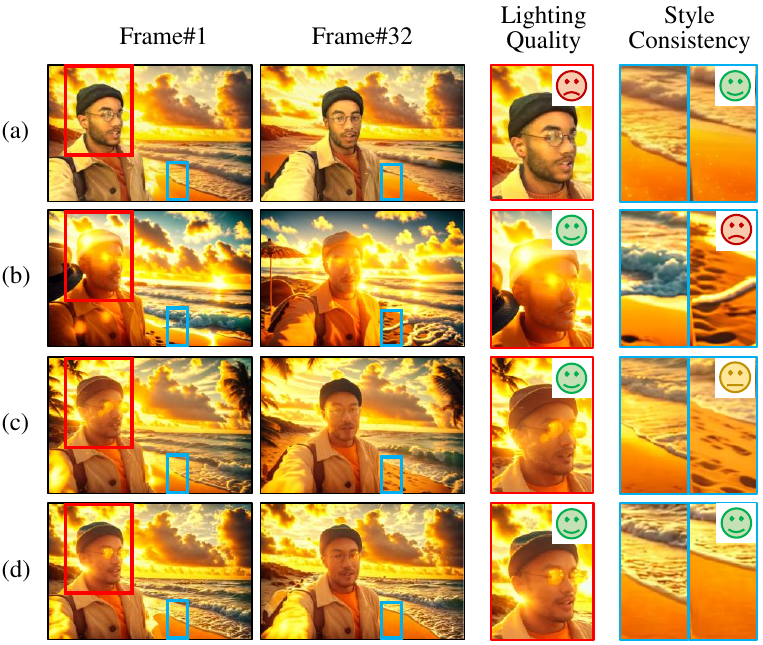}\vspace{-2mm}
\caption{\textbf{Two-step light harmonization.} Our two-step harmonization pipeline with cross-frame attention enables high lighting quality (enlarged red region) and inter-frame style consistency (enlarged blue region). Light harmonization results of (a) $\delta_I$, (b) $\delta_p$, (c) $\delta_I+\delta_p$, (d) $\delta_I+\delta_p+$ cross-frame attention.}\vspace{-5mm}
\label{fig:ic}
\end{figure}


We first extract foreground objects $\mathbf{I}_f$ from $\mathbf{I}$ with 
an image segmentation model BiRefNet~\cite{zheng2024bilateral}.
Now, for each frame of $\mathbf{I}'$, we have a background prompt $p$, a background image $I_b\in\mathbf{I}_b$ and a foreground image $I_f\in\mathbf{I}$. 
We have tried applying both the image-guided and text-guided IC-Light to combine them. However, neither produces reasonable results. 
As shown in Figs.~\ref{fig:ic}(a)(b), the image-guided result $I'=\delta_I(I_f, I_b)$ has an insufficiently harmonized illumination (missing backlight effect), while the text-guided result $I'=\delta_p(I_f, p)$ has a strong illumination effect and suffers from temporal inconsistency and mismatched camera motions in the background due to the lack of image guidance. 

To strike a balance of the illumination effect between $\delta_I$ and $\delta_p$ while simultaneously utilizing the temporally coherent visual guidance from $\mathbf{I}_b$, we propose a two-step harmonization pipeline, as illustrated in Fig.~\ref{fig:pipeline}(b). 
In the first step, we obtain the image-guided result $I'_{img}=\delta_I(I_f, I_b)$. 
In the second step, we take the idea of SDEdit~\cite{meng2021sdedit} to refine the illumination of $I'_{img}$ by denoising it using $\delta_p$. In particular, we add noise of $T_0$ steps ($T_0<T$) to $I'_{img}$ with DDPM forward process~\cite{ho2020denoising}, which is then denoised for $T_0$ steps using $\delta_p$ under the conditions of $I_f$ and $p$. As shown in Fig.~\ref{fig:ic}(c), the foreground lighting is well enhanced. However, such per-frame processing cannot ensure style consistency (\eg, the inconsistent appearance of beaches and the varied position of palm leaves in two frames). 
To alleviate this issue, we employ cross-frame attention~\cite{khachatryan2023text2video,wu2023tune,yang2023rerender} to $\delta_I$ and $\delta_p$. We replace $\delta$'s self-attention layers with cross-frame attention layers, where all frames aggregate key and value features from the first frame rather than themselves. 
As a result, the style consistency is strengthened, as shown in Fig.~\ref{fig:ic}(d).
Note that our pipeline allows one to adjust the illumination effect via $T_0$, \ie, a large $T_0$ produces results with intensive lights and shadows. 

\begin{figure}[t]
\centering
\includegraphics[width=\linewidth]{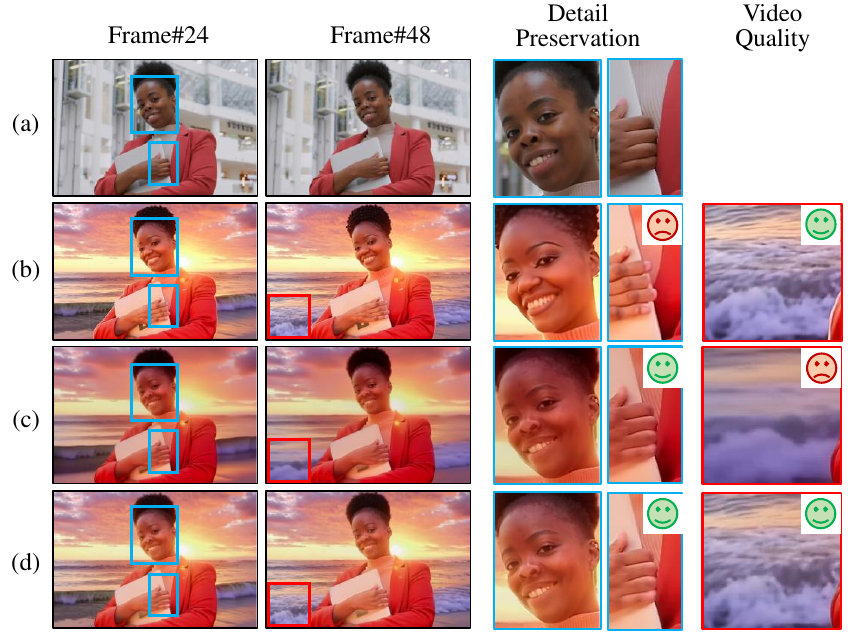}\vspace{-2mm}
\caption{\textbf{Consistency enhancement via Refinement Projection Algorithm (RPA).} (a) Input video. (b) Video diffusion model strengthens temporal consistency, but creates inconsistent foreground appearance. (c) Foreground refinement without RPA introduces quality degradation in the background. (d) RPA effectively constrains foreground details while preserving other regions.}\vspace{-4mm}
\label{fig:rpa}
\end{figure}

\subsubsection{Consistency Enhancement}

The video $\mathbf{I}_L$ generated in Sec.~\ref{sec:method-2} still presents two issues: 1) Even introducing cross-frame attention for global style consistency, there are still pixel-level jitters between frames. 2) $\mathbf{I}_L$'s foreground details do not exactly match $\mathbf{I}$'s foreground details. Therefore, we aim to take advantage of the ability of the video diffusion model $\epsilon_{\theta}$ to improve both the inter-frame temporal continuity and the foreground detail consistency. As with Sec.~\ref{sec:method-2}, our high-level idea is to use SDEdit to refine the temporal consistency of $\mathbf{I}_L$ by denoising it using $\epsilon_{\theta}$ for $T_1$ steps ($T_1 < T$).
We additionally apply edge-based ControlNet to $\epsilon_{\theta}$ to preserve the main structure of $\mathbf{I}$, as shown in Fig.~\ref{fig:pipeline}(c). 
However, ControlNet only provides coarse structure guidance, failing to maintain the identity in Fig.~\ref{fig:rpa}(b). Thus, we propose a \textbf{Refinement Projection Algorithm~(RPA)} to enforce the consistency of the foreground details at the pixel level.

The key idea is to transfer the high-frequency details (since high-frequency information of a frame typically depicts its edges and textures, while low-frequency information characterizes its colors and illuminations) from $\mathbf{I}$ to $\mathbf{I}_L$  in the foreground during SDEdit denoising. As analyzed in Sec.~\ref{sec:intro}, the compact 3D latent space impedes direct high-frequency refinement in the pixel domain. Thus, we first decode the latent back to the pixel domain to apply the refinement, and then encode the refined video back to the latent space. To avoid quality degradation from the inherent reconstruction error of the 3D VAE, RPA computes a zero-error projection direction to guide the encoding. Specifically, RPA has two parts: foreground refinement and DDIM denosing with RPA. 




\textbf{Foreground Refinement}.~To avoid the interference of noises, we follow common practice~\cite{yang2023rerender} to operate on noise-free latent $x_0^t$ in Eq.~(\ref{eq:x0t}). $x_0^t$ is first decoded back to video $\mathbf{I}_0^t=\mathcal{D}(x_0^t)$. 
Subsequently, $\mathbf{I}_0^t$ and $\mathbf{I}$ are decomposed into their 
low-frequency (LF) and high-frequency~(HF) components.
In the foreground region of the refined video $\tilde{\mathbf{I}}_0^t$, we combine $\mathbf{I}_0^t$'s LF component and $\mathbf{I}$'s HF component. 
The background region of $\tilde{\mathbf{I}}_0^t$ is set to the inpainted $\mathbf{I}$, which removes the foreground object using ProPainter~\cite{zhou2023propainter}. The refinement details is summarized in Algorithm~\ref{alg:ref}.


\setlength{\textfloatsep}{1pt}

\begin{algorithm}[t]
\KwIn{Edited video $\mathbf{I}_0^t$, original input video $\mathbf{I}$}
\KwOut{Refined video $\tilde{\mathbf{I}}_0^t=\text{Refine}(\mathbf{I}_0^t, \mathbf{I})$}
\( \mathbf{I}^t_{0,\text{LF}} = \text{GaussianBlur}(\mathbf{I}_0^t) \)\;
\( \mathbf{I}^t_{0,\text{HF}} = \mathbf{I}_0^t - \mathbf{I}^t_{0,\text{LF}} \)\;
\( \mathbf{I}_{\text{LF}} = \text{GaussianBlur}(\mathbf{I}) \)\;
\( \mathbf{I}_{\text{HF}} = \mathbf{I} - \mathbf{I}_{\text{LF}} \)\;
\( \mathbf{M}^t_0 = \text{ForegroundSegmentation}(\mathbf{I}) \)\;
\(\mathbf{I}_{\text{BG}} = \text{Inpaint}(\mathbf{I}_0^t, \text{ForegroundSegmentation}(\mathbf{I}_0^t))\)\;
\( \tilde{\mathbf{I}}_0^t = \mathbf{M}^t_0 \cdot (\mathbf{I}_{\text{HF}} + \mathbf{I}^t_{0,\text{LF}}) + (\mathbf{1} - \mathbf{M}^t_0) \cdot \mathbf{I}_{\text{BG}} \)\; 
\caption{Foreground Refinement}
\label{alg:ref}
\end{algorithm}

\textbf{DDIM Denosing with RPA}. We would like to re-encode the refined video $\tilde{\mathbf{I}}_0^t$ back into latent space as $\hat{x}_0^t$, to replace the original $x_0^t$ during DDIM denoising. 
Ideally, apart from refined HF details, $\hat{x}_0^t$ should remain unchanged compared to $x_0^t$. However, there are two places where errors could be introduced. First, encoding and decoding is not strictly reversible; second, the stochastic nature of VAE causes inevitable discrepancies. Actually, VAE outputs the mean and standard deviation of the latent: $\hat{\mu},\hat{\sigma}=\mathcal{E}(\tilde{\mathbf{I}}_0^t)$, and $\hat{x}_0^t$ is sampled by reparameterization: $\hat{x}_0^t=\hat{\mu}+\epsilon\hat{\sigma}$ with $\epsilon\sim N(0,1)$, which introduces randomness. As the DDIM denoising iterates, such randomness and errors accumulate, resulting in a blurred background as in Fig.~\ref{fig:rpa}(c). 

\begin{figure*}[htp]
    \centering
    \includegraphics[width=0.94\textwidth]{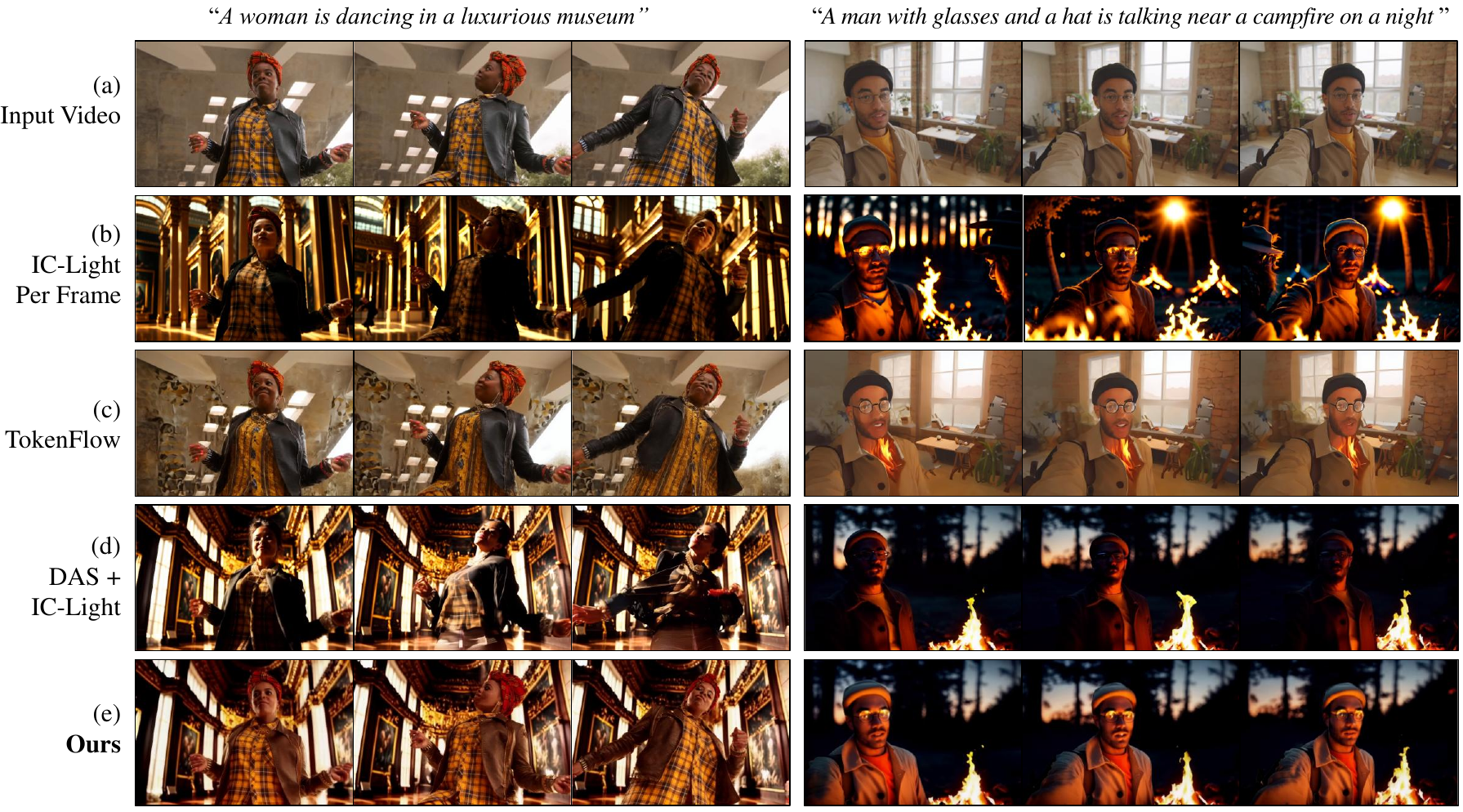}\vspace{-2mm}
    \caption{\textbf{Visual comparison on zero-shot video background replacement.} Full video results are provided in the supplementary material.}\vspace{-4mm}
    \label{fig:compare}
\end{figure*} 

Instead of a random $\epsilon$, our RPA uses a deterministic $\hat{\epsilon}$ that is computed to ensure a perfect reconstruction on $x_0^t$. Specifically, we assume a perfect reconstruction $\mu + \hat{\epsilon}\sigma=x_0^t$. Note that $x_0^t$ and $\mu,\sigma = \mathcal{E}(\mathcal{D}(x_0^t))$ are all available, so we can deterministically calculate $\hat{\epsilon}$.
Then, for the refined video $\tilde{\mathbf{I}}_0^t = \text{Refine}(\mathbf{I}_0^t, \mathbf{I})$, we obtain $\hat{\mu},\hat{\sigma}=\mathcal{E}(\tilde{\mathbf{I}}_0^t)$ and the final projection solution is $\hat{x}_0^t = \hat{\mu} + \hat{\epsilon} \hat{\sigma}$. 
We summarize our proposed RPA in Algorithm~\ref{alg:rpa}.
Our key insight is that if no refinement is applied (\ie, $\tilde{\mathbf{I}}_0^t=\mathbf{I}_0^t$), this projection will cause $\hat{x}_0^t$ exactly equal to $x_0^t$. 
Such an alignment property ensures the resulting video's background area remains almost identical with only foreground details refined, which is also verified by Fig.~\ref{fig:rpa}(d).

\begin{algorithm}[t]
\KwIn{Initial noise \(x_{T_1}\), input video \(\mathbf{I}\), condition $c$}
\KwOut{Refined denoised result \(x_0\)}
\For{$t = T_1 \dots 1$}{
    \( x_0^t = \big(x_t - \sqrt{1 - \alpha_t} \epsilon_{\theta}(x_t, c, t)\big)/\sqrt{\alpha_t} \)\;
    \( \mathbf{I}_0^t = \mathcal{D}(x_0^t) \)\;
    \( \mu, \sigma = \mathcal{E}(\mathbf{I}_0^t) \)\;
    \( \tilde{\mathbf{I}}_0^t = \text{Refine}(\mathbf{I}_0^t, \mathbf{I}) \)\;
    \( \hat{\mu}, \hat{\sigma} = \mathcal{E}(\tilde{\mathbf{I}}_0^t) \)\;
    \( \hat{\epsilon} = (x_0^t - \mu)/ \sigma \)\;
    \( \hat{x}_0^t = \hat{\mu} + \hat{\epsilon} \hat{\sigma} \)\;
    \( x_{t-1} = \sqrt{\alpha_{t-1}} \hat{x}_0^t + \sqrt{1 - \alpha_{t-1}} \epsilon_{\theta}(x_t, c, t) \)\;
}
\caption{DDIM Denoising with RPA}
\label{alg:rpa}
\end{algorithm}

\vspace{-2mm}
\section{Experiments}\vspace{-1mm}
\label{sec:experiments}

\textbf{Implementation Details}.
We instantiate \METHODNAME~with CogVideoX~\cite{yang2024cogvideox} as the video diffusion model $\epsilon_\theta$, and IC-Light~\cite{zhang2025scaling} as the image background replacement model $\delta_p$ and $\delta_I$. We set $T=20$, $(T_0, T_1)$ to $(0.7T, 0.7T)$ and $(0.4T, 0.4T)$ for strong and weak illumination effects, respectively, due to different scenario needs. All experiments are conducted on a single NVIDIA 4090 GPU with CPU offload activated for CogVideoX. 
The testing videos are uniformly resized to 480$\times$720 and trimmed to 49 frames to comply with CogVideoX specifications. Each video requires approximately 12 minutes for inference (which can be further accelerated with CPU offload off if larger GPU memory is available).
The code of this work will be released along with the publication of this paper.

\textbf{Baseline}. 
Since there are few other works exactly handling our zero-shot video background replacement task, we choose the following most related baselines for comparison.
\begin{itemize}
    \item IC-Light~\cite{zhang2025scaling}: A state-of-the-art image background replacement model. We apply it frame-by-frame.
    \item TokenFlow~\cite{geyer2023tokenflow}: A state-of-the-art zero-shot text-guided video editing model.
    \item Diffusion-As-Shader (DAS)~\cite{gu2025diffusion}: A versatile video generation control model. We use its motion transfer function, which creates a new video by transferring motion from an input video to a provided image as the first frame. Here, we use IC-Light to generate the first frame.
\end{itemize}
Note that all above baselines are zero-shot diffusion-based editing methods to ensure a fair comparison.

\textbf{Evaluation}.
We construct a test set consisting of 30 samples and prompts for evaluation, and use the following metrics for evaluation: 
1) Fram-Acc~\cite{qi2023fatezero}: 
The proportion of video frames where the CLIP-based cosine similarity with the target prompt is higher than that with the source prompt, to measure whether the background is successfully edited.
2) Tem-Con~\cite{qi2023fatezero}: CLIP-based cosine similarity between consecutive frames to measure temporal consistency; 
3) ID-Psrv: Preservation of foreground detail of the generated video, measured by the identity loss~\cite{deng2019arcface} between the human face (if applicable) in generated video and the input video; 
4) Mtn-Psrv: Preservation of the motion of the generated video, measured by point motion tracking similarity between the generated video and the input video. We use SpatialTracker~\cite{xiao2024spatialtracker} to track points.

For user study, we invite 24 participants.
Participants are asked to select the best results among the four methods based on three criteria: 1) User-Pmt: how well the result aligns with the prompt, 2) User-Tem: the temporal consistency of the result, 3) User-Psrv: how well the foreground details and motions are preserved, and 4) User-Lgt: the quality of relighting on foreground.

\setlength{\floatsep}{1pt}
\setlength{\textfloatsep}{2pt}

\begin{table}[t]
\begin{center}
\caption{Quantitative comparison and user preference rates}
\resizebox{\linewidth}{!}{
\begin{tabular}{l|ccccc}
\toprule
\textbf{Metric}  & IC-Light & TokenFlow & DAS & Ours\\
\midrule
Fram-Acc $\uparrow$ & \textbf{0.983} & 0.541 & 0.937 & \underline{0.973} \\
Tem-Con $\uparrow$ & 0.945 & 0.981 & \underline{0.986} & \textbf{0.993} \\
ID-Psrv $\downarrow$ &  0.578 & 0.632  & \underline{0.364}  & \textbf{0.313}  \\
Mtn-Psrv $\uparrow$ & 0.844 & \underline{0.985}  & 0.878  & \textbf{0.987} \\
\midrule
User-Pmt &  1.11\% &  1.11\% &  \underline{29.72\%} &  \textbf{68.06\%} \\
User-Tem &  0.56\% &  5.56\% &  \underline{28.61\%} &  \textbf{65.28\%}  \\
User-Psrv &  2.78\% &  \underline{18.33\%} &  17.22\% &  \textbf{61.67\%} \\
User-Lgt &  11.11\% &  11.11\% &  \underline{30.56\%} &  \textbf{47.22\%} \\
\bottomrule
\end{tabular}}
\label{tb:quantitative_evaluation}
\end{center}
\end{table}

\subsection{Comparison to State-of-the-Art Methods}


Figure~\ref{fig:compare} visually compares the proposed method with other baselines. 
IC-light~\cite{zhang2025scaling}, being fundamentally an image diffusion model, inherently suffers from temporal inconsistency. Moreover, it tends to overly relight the subject, even changing the intrinsic properties like the color of the clothes and the headscarf. TokenFlow~\cite{geyer2023tokenflow} demonstrates limited editing capabilities and insufficient foreground detail control, while DAS~\cite{gu2025diffusion} fails to maintain control over foreground motion dynamics and intrinsic appearance properties. In contrast, our method achieves high-quality background replacement and foreground relighting while ensuring temporal consistency and foreground detail consistency. Full results are provided in the supplementary material.

Table~\ref{tb:quantitative_evaluation} gives quantitative evaluations. IC-Light achieves the best Fram-Acc as it is specifically trained for this background replacement task, without the need to consider temporal consistency. Our method achieves the second-best Fram-Acc, and the best results across all other metrics and user preferences, striking a good balance between single-frame relighting quality and overall video smoothness. 

\subsection{Ablation Study}

\begin{figure}[t]
    \centering
    \includegraphics[width=\linewidth]{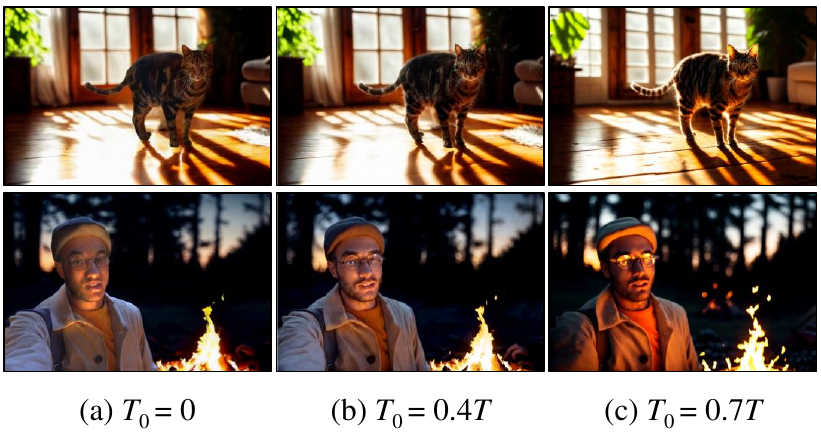}\vspace{-2mm}
    \caption{\textbf{Effect of $T_0$ in illumination harmonization.} With increased $T_0$, $\delta_p$ assumes a more prominent role in facilitating intensive light and shadow effect for foreground.}\vspace{2mm}
    \label{fig:compare2}
\end{figure} 

\begin{figure}[t]
    \centering
    \includegraphics[width=\linewidth]{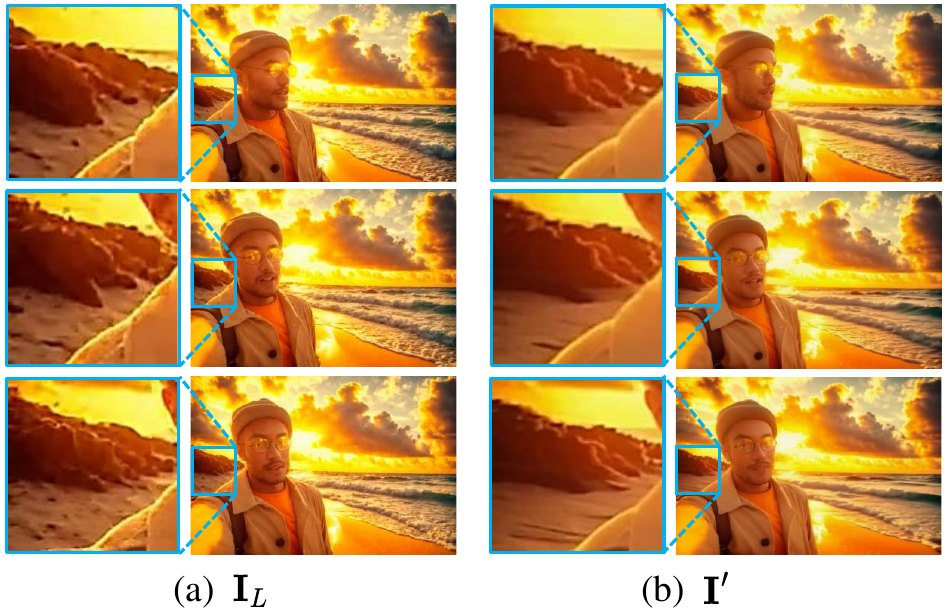}\vspace{-2mm}
    \caption{\textbf{Effect of Consistency Enhancement.} The temporal prior of video diffusion model effectively helps improve temporal consistency, \eg, eliminating the inconsistent appearance of rocks as in the enlarged blue region.}
    \label{fig:compare3}
\end{figure} 

To validate the contributions of different modules to the overall performance, we systematically deactivate specific modules in our framework. The results are reported in Figs.~\ref{fig:ic}, \ref{fig:rpa}, \ref{fig:compare2}, \ref{fig:compare3} and Table~\ref{tb:abla}.
\begin{itemize}
    \item \textbf{Two-Step Harmonization}. In the Image Harmonization stage, we employ IC-Light $\delta_I$ and $\delta_p$ to provide improved illumination for videos, enabling better integration of the foreground into the background. A naive one-step harmonization (\ie, w/o $\delta_p$ or  $T_0=0$) would result in video models generating foregrounds with less natural lighting, as reported in Table~\ref{tb:abla}. As $T_0$ increases, $\delta_p$ gradually plays a stronger role in achieving more natural foreground lighting, as shown in Fig.~\ref{fig:compare2}.
    \item \textbf{Cross-Frame Attention}. The effect of cross-frame attention injection is studied in Fig.~\ref{fig:ic}. Disabling cross-frame attention leads to severe inter-frame appearance discrepancy in the generated results (\eg, footprints suddenly appeared on the beach), degrading temporal consistency.
    \item \textbf{Temporal Prior}. The Consistency Enhancement (Cst-Enh) stage optimizes the foreground details and overall temporal consistency of the video $\mathbf{I}_L$ generated in the second stage. 
    Without Cst-Enh, Tem-Con drops significantly as in Table~\ref{tb:abla}.
    As demonstrated in Fig.~\ref{fig:compare3}, $\mathbf{I}_L$ suffers from inconsistent rock in the background. By leveraging the strong prior of video diffusion models, this issue is effectively solved in the stage-three result $\mathbf{I}'$.
    \item \textbf{RPA}.  RPA performs high-frequency detail refinement on the latent. Without RPA, the identity discrepancy becomes greater as in Table~\ref{tb:abla} and Fig.~\ref{fig:rpa}(b). 
    A naive high-frequency detail refinement through decoding and encoding leads to a blurry background, as in Fig.~\ref{fig:rpa}(c). Our designed RPA provides a deterministic sampling scheme that well preserves the non-refined regions like background areas, as in Fig.~\ref{fig:rpa}(d).    
\end{itemize}


\section{More Results}

\begin{figure}[t]
    \centering
    \includegraphics[width=\linewidth]{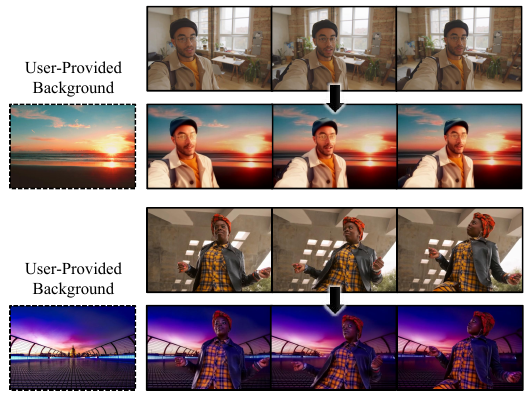}\vspace{-2mm}
    \caption{\textbf{Image-guided background replacement.} In addition to text prompts, our method also supports replacing the background based on a user-provided image.}\vspace{1mm}
    \label{fig:ext}
\end{figure}

\textbf{Image-Guided Video Background Replacement}.
Our method can be easily adapted to image prompts. In the first stage, we generate the first frame with $\delta_I$ and a user-provided background scene image, while the subsequent stages remain identical to those with text prompts. Our image-guided results are shown in Fig.~\ref{fig:teaser} and Fig.~\ref{fig:ext}.

\begin{figure}[t]
    \centering
    \includegraphics[width=0.76\linewidth]{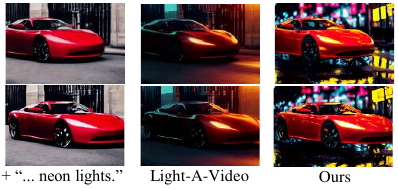}\vspace{-2mm}
    \caption{\textbf{Comparison to Light-A-Video.} As concurrent work, Light-A-Video only relights backgrounds instead of generating new content.}\vspace{3mm}
    \label{fig:lav}
\end{figure}

\textbf{Comparison to Light-A-Video}.
We further provide a visual comparison with a concurrent work of Light-A-Video~\cite{zhou2025light} in Fig.~\ref{fig:lav}. The two methods, both based on CogVideoX, produce outputs of comparable quality. However, the CogVideoX implementation of Light-A-Video can only relight the existing background, while our method can generate new background content.

\begin{table}[t]
\begin{center}
\caption{Quantitative ablation study}\vspace{-1mm}
\resizebox{\linewidth}{!}{
\begin{tabular}{l|ccccc}
\toprule
\textbf{Metric}  & w/o $\delta_p$ & w/o Cst-Enh & w/o RPA &Full \\
\midrule
Fram-Acc $\uparrow$ & 0.966 & \underline{0.970} & \underline{0.970} & \textbf{0.973} \\
Tem-Con $\uparrow$ & \underline{0.989} & 0.961 & 0.987 & \textbf{0.993} \\
ID-Psrv $\downarrow$ &  \underline{0.329} & 0.353  & 0.371  & \textbf{0.313}  \\
Mtn-Psrv $\uparrow$ & \textbf{0.987} & 0.973  & \underline{0.984}  & \textbf{0.987} \\
\bottomrule
\end{tabular}}
\label{tb:abla}
\end{center}
\end{table}

\section{Limitations}

\begin{figure}[t]
    \centering
    \includegraphics[width=\linewidth]{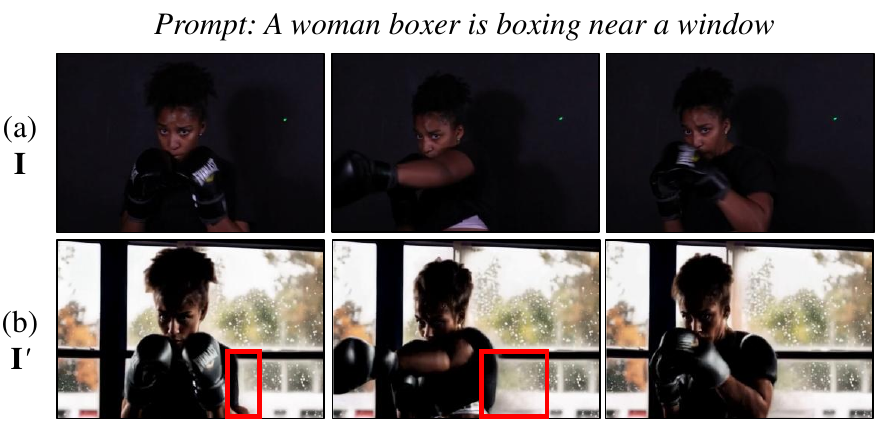}\vspace{-2mm}
    \caption{\textbf{Limitations.}~Low-quality input, poor boundary condition and fast movement degrades the performance of our method. Red boxes indicate the mismatched and blurry boundary region.}\vspace{2mm}
    \label{fig:fail}
\end{figure}

While \METHODNAME~demonstrates promising results, several limitations remain. Figure~\ref{fig:fail} gives a typical example. 
1) Low-quality inputs (\eg, low-resolution/blurry) reduce high-frequency detail transfer, causing blurred results like Fig.~\ref{fig:fail}'s hair; 
2) Unclear foreground-background boundaries lead to mismatched inpainting and enlarged blurry regions around subjects; 
3) Rapid motion challenges diffusion models, causing artifacts on the left arm.


\section{Conclusion and Discussion}

In this paper, we propose \METHODNAME, a zero-shot framework for video background replacement and foreground relighting that achieves high temporal consistency and detail fidelity without task-specific training. 
Specifically, by integrating motion-aware video diffusion for background generation, extending image relighting models with cross-frame attention, and introducing the Refinement Projection Algorithm to preserve foreground details in latent space, our method outperforms existing approaches in both lighting harmonization and temporal coherence. 

One possible future direction is to investigate the extension of diverse editing tasks (\eg, recolorization, stylization, facial attribute editing, inpainting) to video domains with the temporal prior of large video diffusion models. 

\small{
\noindent
\textbf{Acknowledgments.} This work was supported in part by the National Natural Science Foundation of China under Grant 62471009, in part by CCF-Tencent Rhino-Bird Open Research Fund, in part by the Key Laboratory of Science, Technology and Standard in Press Industry (Key Laboratory of Intelligent Press Media Technology), and in part by The Fundamental Research Funds for the Central Universities, Peking University.
}

{
    \small
    \bibliographystyle{ieeenat_fullname}
    \bibliography{main}
}

\end{document}